\pdfoutput=1

\documentclass[11pt]{article}

\usepackage{authblk}
\usepackage[final]{acl}
\usepackage{relsize}
\usepackage{times}
\usepackage{latexsym}
\usepackage{tabularx}
\usepackage{booktabs} 
\usepackage[T1]{fontenc}

\usepackage[utf8]{inputenc}

\usepackage{microtype}

\usepackage{inconsolata}
\usepackage{amsmath}
\usepackage{amsfonts}
\usepackage{footnote}
\usepackage{booktabs}
\usepackage{hyperref} 
\usepackage{multirow}
\usepackage{colortbl}
\usepackage{xcolor}
\usepackage{subfigure}

\usepackage{graphicx}
\usepackage{makecell}

\begin{document}

%
%

\title{Text Annotation via Inductive Coding: Comparing Human Experts to LLMs in Qualitative Data Analysis}



\author[1,2]{\textbf{Angelina Parfenova}}
\author[1]{\textbf{Andreas Marfurt}}
\author[1]{\textbf{Alexander Denzler}}
\author[2]{\\\textbf{Juergen Pfeffer}}
\affil[1]{Lucerne University of Applied Sciences and Arts}
\affil[2]{Technical University of Munich}
\maketitle
\renewcommand{\thefootnote}{\fnsymbol{footnote}}
\begin{abstract}

This paper investigates the automation of qualitative data analysis, focusing on inductive coding using large language models (LLMs). Unlike traditional approaches that rely on deductive methods with predefined labels, this research investigates the inductive process where labels emerge from the data. The study evaluates the performance of six open-source LLMs compared to human experts. As part of the evaluation, experts rated the perceived difficulty of the quotes they coded. The results reveal a peculiar dichotomy: human coders consistently perform well when labeling complex sentences but struggle with simpler ones, while LLMs exhibit the opposite trend. Additionally, the study explores systematic deviations in both human and LLM generated labels by comparing them to the golden standard from the test set. While human annotations may sometimes differ from the golden standard, they are often rated more favorably by other humans. In contrast, some LLMs demonstrate closer alignment with the true labels but receive lower evaluations from experts.


\end{abstract}

\section{Introduction}

Qualitative data analysis (QDA) is an important research method across various fields such as marketing, media studies, social science, psychology, medical research, and others \cite{Avjyan2005,  Creswell2016, mohajan2018qualitative, Flick2018, Leeson2019, brennen2021qualitative}. Unlike quantitative research, which relies on numerical data and statistical analysis, qualitative research captures the richness and complexity of human experiences, behaviors, and social phenomena \cite{Denzin2005, Patton2014}. It explores research questions in more details, providing insights that are often missed by quantitative methods. Yet, this depth of understanding comes at a cost—QDA is naturally labor-intensive, requiring thorough manual work that are both time-consuming and sensitive to inconsistencies and subjective biases \cite{Morse2015, bumbuc2016subjectivity}.

\begin{figure}
    \centering
    \includegraphics[width=0.8\linewidth]{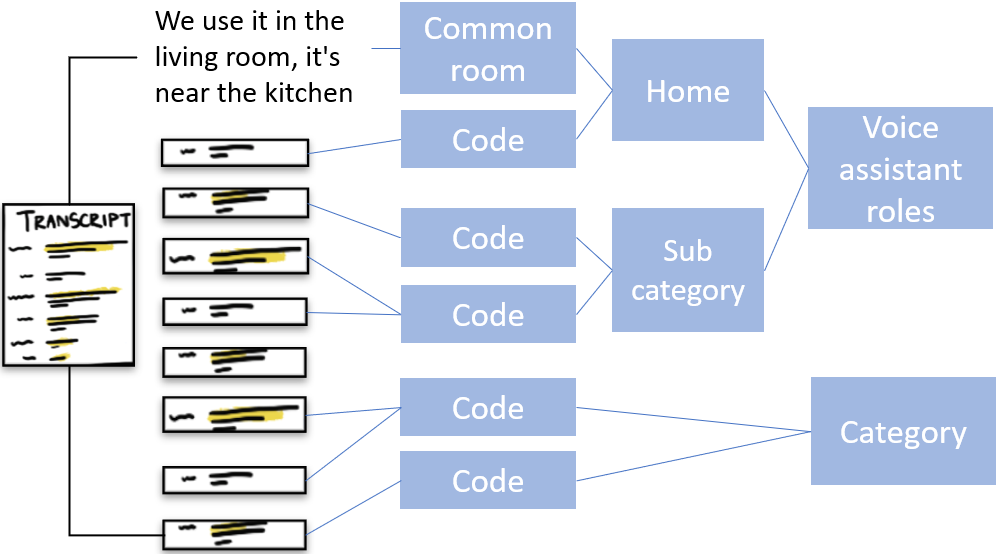}
    \caption{Coding in \textit{thematic analysis}. The source text is split into quotes. The main idea of a paragraph is extracted and becomes a \textit{code} (open coding). Then, this list of codes is hierarchically grouped into more abstract categories (axial coding).}
    \label{fig:coding}
\end{figure}

One of the most critical and demanding stages of QDA, specifically of \textit{thematic analysis}, is the process of \textit{coding}. Coding involves the systematic identification and labeling of significant themes, ideas, attitudes, and topics within a body of text \cite{Charmaz2014, glaser2017discovery}. This method consists of two stages: open coding and axial coding (see Figure \ref{fig:coding}) which aim to summarize the main ideas from sentences into \textit{codes} and then categorize them, establishing hierarchy \cite{Saldana2016}. Despite its importance, coding is time-consuming, often taking weeks to complete for large datasets \cite{Alshenqeeti2014, Hennink2020}. Moreover, the manual nature of this process makes it prone to subjective interpretation, which can lead to variability in the results \cite{Ryan2003, MacQueen2008}.

Automating QDA is increasingly important, as traditional methods like Topic Modeling and Wordnet hierarchies capture keywords but often miss deeper insights \cite{Leeson2019, Parfenova_Lucerne_2024}. Advances in NLP, especially LLMs, offer potential for reducing manual effort in coding, though their ability to match human analysis remains uncertain. This paper explores how LLMs can automate the open coding process, comparing their performance to human experts through experiments in zero-shot, few-shot, and fine-tuning scenarios. A key finding of this study is that fine-tuning with as few as 100 examples can achieve sufficient performance, which is particularly beneficial for computational social science research, where data collection remains a challenge \cite{lazer2020computational}.

\section{Related Work}



Qualitative coding involves the systematic categorization of textual data to identify patterns, themes, and insights. In this process, each significant statement or segment of text is assigned a \textit{code} that encapsulates its core idea. According to the definition by \citet{Saldana2016}, a code is "often a word or short phrase that symbolically assigns a summative, salient, essence-capturing, and/or evocative attribute for a portion of language-based or visual data." In one of the most popular methods of QDA, \textit{thematic analysis}, once these segments are coded, they are grouped into broader categories that highlight underlying hierarchy between codes. The data itself can consist of interviews, documents, field notes, or any other source of qualitative information. To explain the process more simply, we first summarize the main idea of each quote (sentence or paragraph). Then, we group these summaries into larger categories. This involves examining all the ideas we've identified and determining how they fit together into broader themes.

\paragraph{Methods similar to coding}



One of the most extensively studied approaches is the use of topic modeling and word embeddings. These techniques are often compared directly to traditional open coding methods to evaluate their effectiveness. For instance \citet{Leeson2019} used Latent Dirichlet Allocation (LDA) by  \citet{blei2003latent} to extract topics from text data, assigning weights to words that represent the identified topics. In this, the words in topics were compared to the codes created by human coders. A more recent development in this area involves using BERT embeddings with \textit{hierarchical density-based spatial clustering of applications with noise (HDBSCAN)} \cite{grootendorst2022bertopicneuraltopicmodeling, Parfenova_Lucerne_2024}. This method provides a more detailed and contextually aware representation of the data. However, this technique still extracts only existing words from the text rather than generating new ideas based on the context.

Another approach explored in the literature involves leveraging WordNet \cite{miller1990introduction}, a lexical database that represents the semantic relationships between words in a hierarchical structure \cite{10.1016/j.eswa.2014.10.023, Guetterman2018}. However Wordnet has limited lexical coverage and is not actuvely maintained.  ConceptNet, on the other hand, extends beyond WordNet by capturing common sense knowledge and broader connections between concepts, making it more suitable for qualitative coding \cite{liu2004conceptnet}, but still using only words present in the text, instead of generating new ones.




As for the automation of coding using LLMs, it is worth mentioning that there are two primary approaches: deductive and inductive. Deductive coding is theory-driven, where predefined codes are applied to the data. In contrast, inductive coding is data-driven, allowing codes to emerge organically from the data. Some studies, such as \citet{Xiao_2023, piano-etal-2023-qualitative, matter2024close, ziems2024can, fischer-biemann-2024-exploring}, have explored the use of LLMs for automatic deductive code generation where labels are predefined. However, our approach utilizes an inductive coding method based on \textit{grounded theory} \cite{glaser2017discovery}, allowing insights to naturally emerge from the data.

\paragraph{Existing software for coding}

Qualitative researchers usually utilize specialized software such as Atlas.ti\footnote{\url{https://atlasti.com/} }, Dedoose\footnote{\url{https://www.dedoose.com/} }, MAXQDA\footnote{\url{https://www.maxqda.com/} } to aid in manual coding. These tools provide a user-friendly interface for tagging, categorizing, and organizing data. While these platforms offer significant convenience and streamline the workflow, they do not perform coding itself. Instead, they serve as digital extensions of traditional qualitative methods, making it easier to manage large volumes of data.

\section{Dataset}

\begin{figure}
    \centering
    \includegraphics[width=\linewidth]{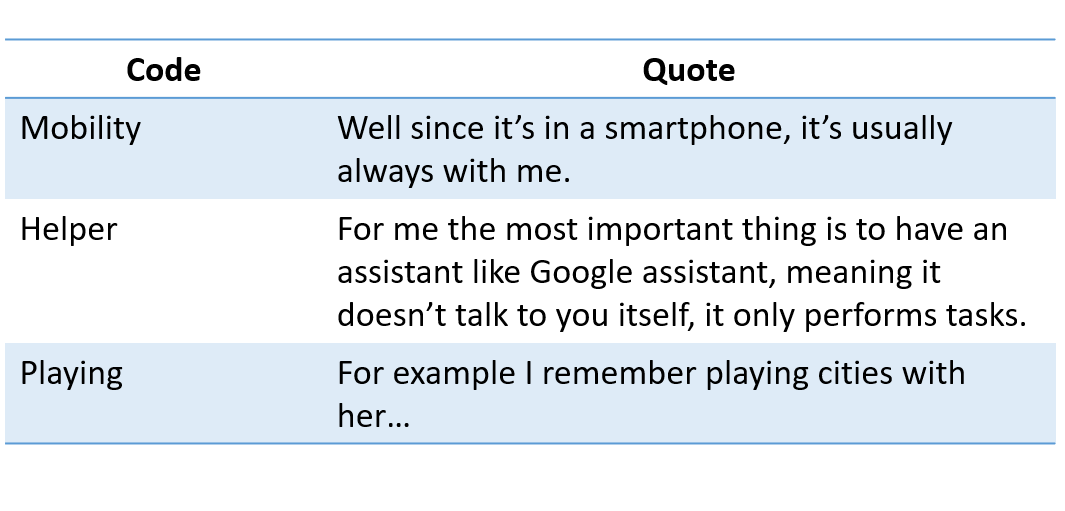}
    \caption{Dataset examples}
    \label{fig:dataset}
\end{figure}

The dataset was compiled from student and professor contributions across three social science faculties of different universities. It consists of 600 \textit{code}-\textit{quote} pairs (see example in Figure \ref{fig:dataset}). As shown in Table \ref{tab:research_summary}, most of these studies were based on interviews covering various topics such as values, social expectations, interaction with technology, while one of them involved the analysis of online reviews. The \textit{Code} column in the dataset represents the \textit{golden standard}, established by consensus among 3 to 5 coders. Coders initially labeled quotes independently, then discussed and agreed on the final golden standard label.

To enhance the dataset, an additional 400 \textit{code}-\textit{quote} pairs were incorporated from the SemEval-2014 dataset Task 4, which consists of online reviews \cite{pontiki-etal-2014-semeval}. This data was manually coded by sociologists, who extracted the main idea of each review. Experiments demonstrated that models trained on the augmented dataset outperformed those trained on the original dataset (see Table \ref{tab:augmentation}). As a result, all subsequent experiments were conducted using the augmented dataset of 1,000 examples. The test set size was set to 100 examples (see Table \ref{tab:dataset_summary}). The dataset was split into training and testing sets without a separate validation set. Hyperparameters were selected based on the training results and evaluated on the test set.

\begin{table}[ht]
\centering
\tiny
\begin{tabular}{@{}lll@{}}

\toprule
\textbf{N Quotes} & \textbf{Description} \\ \midrule
\multicolumn{2}{c}{\textbf{Social Science Studies Data: 600 quotes}} \\ \midrule
     78                  & Study about interaction with self-tracking devices (interviews) \\ 
 22                  & Study about life transitions and mobility  (interviews)          \\ 
 82                  & Study about interaction with voice assistants (interviews)           \\ 
 28                  & Study about museums and cultural experiences (interviews)        \\ 
 25                  & Study on doctors' experiences with pregnant women (interviews)    \\ 
 110                 & Study on universal and national values (interviews)        \\ 
 24                  & Study on procrastination and budget planning (interviews)              \\ 
 56                  & Study on technology interactions and user feedback (reviews)  \\ 
 175                 & Study about social expectations (interviews)   \\ \midrule \multicolumn{2}{c}{\textbf{SemEval 2014; Task 4: 400 quotes}} \\ \midrule
211                 & Restaurant reviews      \\ 
 189                 & Laptop reviews     \\ \bottomrule
\end{tabular}
\caption{Summary of Data Sources with descriptions.}
\label{tab:research_summary}
\end{table}


\begin{table}[h!]
\centering
\tiny
\begin{tabular}{l c c c}
\toprule
\textbf{Statistic} & \textbf{Overall} & \textbf{Train} & \textbf{Test} \\
\midrule
Total Quotes & $1000$ & $900$ & $100$ \\
Social Science Data & $600$ & $550$ & $50$ \\
SemEval Data & $400$ & $350$ & $50$ \\
Num of Data Sources & $11$ & $11$ & $11$ \\
Unique Codes & $680$ & $624$ & $94$ \\
Average Quote Length & $254.75_{274.28}$ & $280.89_{280.89}$ & $234.80_{201.61}$ \\
Average Code Length & $19.95_{10.43}$ & $20.04_{10.70}$ & $19.27_{10.53}$ \\
\bottomrule
\end{tabular}
\caption{Summary statistics of the dataset and train/test splits. Subscript refers to standard deviation where applicable.}
\label{tab:dataset_summary}
\end{table}

\begin{table*}[htbp]
    \centering
    \tiny  
    \begin{tabular}{lccccccccccc}
        \toprule
        \textbf{Dataset Size} & \textbf{Parameters} & \multicolumn{3}{c}{\textbf{BERTScore}} & \multicolumn{3}{c}{\textbf{ROUGE}} \\
     &  & $P_{std}$ & $R_{std}$ & $F1_{std}$ & $1$ & $2$ & $L$ \\
        \midrule        
        \textbf{900 with augmentation} & & & & & & & & & &  \\
        Llama3 (instruct) & $8$B & $0.713_{0.060}$ & $0.787_{0.084}$ & $0.747_{0.062}$ & $0.142$ & $0.033$ & $0.136$  \\
        Falcon (instruct) & $7$B & $\mathbf{0.744}_{0.100}$ & $0.788_{0.100}$ & $\mathbf{0.764}_{0.096}$ & $\mathbf{0.204}$ & $\mathbf{0.089}$ & $\mathbf{0.202}$  \\
        Mistral (instruct) & $7$B & $0.728_{0.076}$ & $\mathbf{0.790}_{0.094}$ & $0.756_{0.078}$ & $0.175$ & $0.075$ & $0.166$  \\
        Vicuna (instruct) & $7$B & $0.726_{0.081}$ & $0.788_{0.095}$ & $0.754_{0.080}$ & $0.188$ & $0.077$ & $0.185$  \\
        Gemma (instruct) & $7$B & $0.721_{0.083}$ & $0.775_{0.092}$ & $0.746_{0.081}$ & $0.165$ & $0.059$ & $0.159$  \\
        Tinyllama (chat) & $1.1$B & $0.738_{0.091}$ & $0.781_{0.095}$ & $0.758_{0.088}$ & $0.185$ & $0.073$ & $0.179$  \\
        \midrule
        \textbf{500 without augmentation} & & & & & & & &  & & \\
        Llama3 (instruct) & $8$B & $0.714_{0.069}$ & $0.763_{0.078}$ & $0.737_{0.068}$ & $0.137$ & $0.053$ & $0.135$ \\
        Falcon (instruct) & $7$B & $\mathbf{0.735}_{0.098}$ & $0.757_{0.096}$ & $0.745_{0.092}$ & $0.147$ & $0.041$ & $0.146$ \\
        Mistral (instruct) & $7$B & $0.731_{0.095}$ & $\mathbf{0.776}_{0.093}$ & $\mathbf{0.751}_{0.088}$ & $0.180$ & $\mathbf{0.074}$ & $0.173$  \\
        Vicuna (instruct) & $7$B & $0.722_{0.078}$ & $0.763_{0.080}$ & $0.741_{0.074}$ & $0.141$ & $0.039$ & $0.137$  \\
        Gemma (instruct) & $7$B & $0.702_{0.084}$ & $0.769_{0.091}$ & $0.733_{0.081}$ & $0.157$ & $0.068$ & $0.154$ \\
        Tinyllama (chat) & $1.1$B & $0.726_{0.078}$ & $0.773_{0.089}$ & $0.748_{0.077}$ & $\mathbf{0.187}$ & $0.074$ & $\mathbf{0.178}$  \\
        \bottomrule
    \end{tabular}%
        \caption{Model Performance on open coding task with and without augmentation. The prompt used was 'Summarize the main idea of a sentence.'}
    \label{tab:augmentation}
\end{table*}

\section{Automatic evaluation}

As previously mentioned, this study focuses exclusively on the open coding phase, while categorizing and clustering codes into higher-order categories (axial coding) remains a separate task that requires distinct experimentation and evaluation, and will be covered in a future work. According to established guidelines, open coding does not necessitate prior knowledge of the research topic, whereas axial coding heavily relies on such knowledge \cite{miles1994qualitative, glaser2017discovery}.

In this study, we compared several open source models: Llama3 \cite{touvron2023llama}, Falcon \cite{pineda2023falcon}, Mistral \cite{mistral2023}, Vicuna \cite{vicuna2023}, Gemma \cite{gemma2024}, and TinyLlama \cite{tinyllama2023} (see Appendix F), to evaluate their performance in the open coding task. We experimented with different approaches including zero-shot, few-shot (providing 1 to 5 examples), and parameter-efficient fine-tuning \cite{han2024parameterefficientfinetuninglargemodels} using Low-Rank Adaptation \cite{DBLP:journals/corr/abs-2106-09685}.  

\subsection{Metrics}

To evaluate the performance of chosen models, two metrics were employed to capture both lexical and semantic similarity: ROUGE \cite{lin-2004-rouge} and BERTScore \cite{bertscore}. BERTScore is a metric that computes the similarity between BERT token embeddings of two codes, which helps assess the meaning in the generated output compared to the reference. ROUGE is a lexical similarity measure that calculates the overlap of n-grams (1-unigram overlap, 2-bigram overlap, L-longest common subsequence) between the generated text and the reference text. ROUGE is particularly effective in summarization task \cite{fabbri2021summevalreevaluatingsummarizationevaluation}, which is valuable when the exact wording of the output needs to match the reference.


\subsection{Results}

\paragraph{Finetuning}

Results show that Falcon and Mistral consistently performed better than other models across both the BERTScore and ROUGE metrics, particularly when fine-tuned on the augmented dataset. Falcon achieved the highest BERTSscore (0.7642) when trained on the full dataset, suggesting that it is better at capturing the nuances of sentence meaning compared to other models (see Table \ref{tab:augmentation}). Mistral also demonstrated strong performance, especially in its consistency across different dataset sizes, showing a more stable performance with varying data availability.

\begin{figure*}
    \centering
    \includegraphics[width=0.75\linewidth]{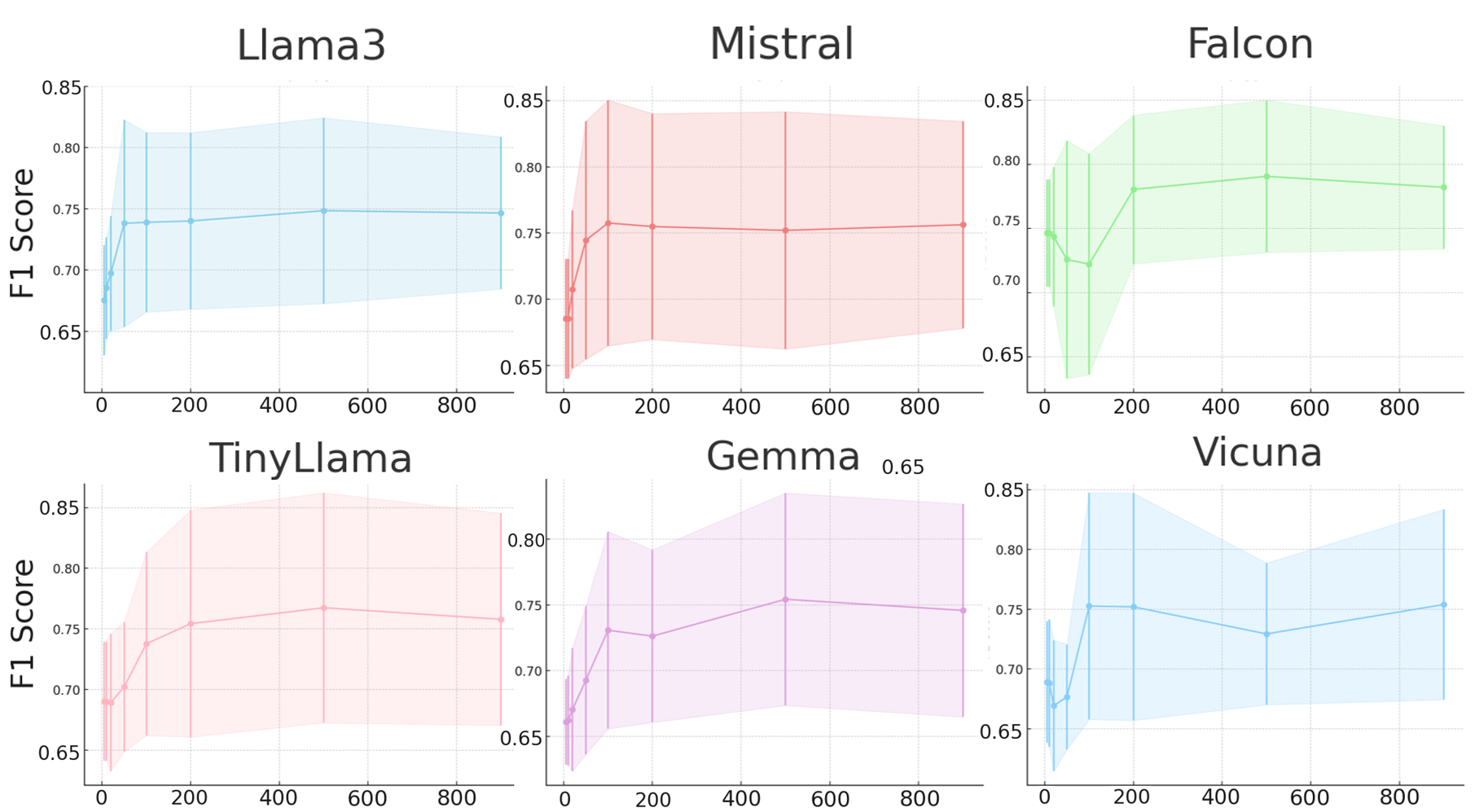}
    \caption{BERT F1 score with an increase of dataset size for all models. The shaded areas represent the standard deviation. The analysis shows how each model benefits from additional data, with some models like Mistral and Falcon displaying higher stability and faster performance gains compared to others. This figure illustrates that the few examples is enough for sufficient finetuning performance.}
    \label{fig:datasizesall}
\end{figure*}

\paragraph{Augmentation}
When comparing results between the augmented dataset (1000 examples) and the smaller dataset (600 examples), it is clear that increasing the training dataset size significantly improves model performance. For instance, Falcon's BERTScore increased from 0.7348 to 0.7642, and Mistral's BERTScore improved from 0.7308 to 0.7562. The results show that all models generally improved in performance as the dataset size increased, supporting that larger training datasets lead to better generalization. However, the most significant finding is that the performance, as measured by the BERTScore, plateaued after approximately 100 examples (demonstrated in the Figure \ref{fig:datasizesall}). This suggests that while additional data beyond 100 examples can still contribute to slight improvements, the majority of performance gains can be achieved with a relatively small amount of data.

\paragraph{Zero-shot and Few-shot}

In this experiment, we evaluated various models across different settings: zero-shot, one-shot, three-shot, and five-shot scenarios. In the zero-shot setting, no examples were provided to the models, and they had to generate codes based solely on the initial prompt. In the one-shot, three-shot, and five-shot settings, the models were given one, three, and five examples, respectively, to help guide their coding (see Table \ref{tab:zeroshot-fewshot}).

\begin{table}[h!]
    \centering
    \tiny
    \begin{tabular}{lrrrr}
        \toprule
        \textbf{Model} & \textbf{Zero-shot} & \textbf{1-shot} & \textbf{3-shot} & \textbf{5-shot} \\
        \midrule
        Llama3        & $0.6713$ & $0.7488$ & $0.7308$ & $0.7473$ \\
        Falcon        & $\boldsymbol{0.7112}$ & $0.7092$ & $0.7195$ & $0.7019$ \\
        Mistral       & $0.6945$ & $\boldsymbol{0.7501}$ & $\boldsymbol{0.7536}$ & $\boldsymbol{0.7613}$ \\
        Vicuna        & $0.6951$ & $0.7496$ & $0.6790$ & $0.6893$ \\
        Gemma         & $0.6951$ & $0.7414$ & $0.7227$ & $0.7339$ \\
        TinyLlama     & $0.6928$ & $0.7444$ & $0.7295$ & $0.6893$ \\
        \bottomrule
    \end{tabular}
    \caption{BERT F1 scores for Zero-shot and Few-shot performance across models}
    \label{tab:zeroshot-fewshot}
\end{table}

The BERTScores across the different models varied depending on the number of examples provided. The performance generally improved when moving from zero-shot to one-shot scenarios, with most models achieving their highest scores with just one example. However, the models exhibited varying behaviors as more examples were provided. Notably, as depicted in Figure \ref{fig:fewshot}, Mistral demonstrated continuous improvement across all scenarios, achieving the highest BERTScore in the five-shot scenario. The best settings of models are demonstrated in Table \ref{tab:main-results}.

\section{Human Expert Evaluation}

The efficacy of LLMs in automating the open coding phase was evaluated through a comparison with human coders. Three expert qualitative researchers with social science educational background manually coded a selection of sentences, and their codes were compared with those produced by six LLMs in their best performance scenarios (see Table \ref{tab:main-results}). The evaluation process was conducted in two stages.

\subsection{Stage 1: Coding and Difficulty Rating}
In the first stage, the human coders participated in an expert coding task, where they were presented with a set of 15 sentences (see Appendix D). According to the definition of a code by \citet{Saldana2016} the coders were asked to generate an open code for each sentence that best encapsulated its core meaning. Each code had to be a word or short phrase summarizing the key idea of the sentence. This open coding process was conducted without any prior knowledge of the golden standard labels.

Additionally, the coders were asked to rate the subjective difficulty of coding each sentence. For each sentence, they chose one of three levels, based on the following criteria:
\textit{Easy (1)} - The sentence is straightforward and the code is obvious;
\textit{Medium (2)} - The sentence requires more thought, but a clear code can still be assigned;
\textit{Difficult (3)} - The sentence is complex or ambiguous, making it difficult to assign a suitable code. In the evaluation process, difficulty was initially assumed based on a 1-to-3 scale, and then compared to the results given by experts. After collecting the difficulty ratings from all coders, the average difficulty was computed across these values.

\begin{table*}[htbp]
    \centering
    \tiny
    \begin{tabular}{lcccccccccc}
        \toprule
        \textbf{Model} & \textbf{Parameters} & \textbf{Adaptation} & \textbf{Prompt} & \multicolumn{3}{c}{\textbf{BERTScore}} & \multicolumn{3}{c}{\textbf{ROUGE}} \\
        \cmidrule(lr){5-7} \cmidrule(lr){8-10}
         &  &  &  & \textit{P} & \textit{R} & \textit{F1} & \textit{1} & \textit{2} & \textit{L} \\
        \midrule
        Llama3 (instruct) & 8B & Finetuning & \raggedright Summarize the main idea of a sentence. & 0.719 & 0.788 & 0.751 & 0.182 & 0.059 & 0.167 \\
        Falcon (instruct) & 7B & Finetuning & \makecell{From the perspective of a social scientist,\\summarize the following sentence as you would\\ in thematic coding.} & 0.745 & 0.792 & 0.766 & 0.210 & 0.089 & 0.211 \\
        Mistral (instruct) & 7B & Finetuning & \makecell{Can you tell me what the main idea of this\\ sentence is in just a few words?} & 0.742 & 0.795 & 0.766 & 0.246 & 0.106 & 0.235 \\
        Vicuna (instruct) & 7B & Finetuning & Summarize the main idea of a sentence. & 0.734 & 0.787 & 0.759 & 0.194 & 0.068 & 0.185 \\
        Gemma (instruct) & 7B & Finetuning & \makecell{If you were a social scientist doing thematic\\ analysis, what code would you give to this citation?} & 0.724 & 0.784 & 0.751 & 0.170 & 0.066 & 0.168 \\
        TinyLlama (chat) & 1.1B & Few-shot (5 examples) & \makecell{Summarize the main idea of a sentence.\\Here are examples:} & 0.768 & 0.744 & 0.755 & 0.176 & 0.026 & 0.176 \\
        \bottomrule
    \end{tabular}
    \caption{Performance of various open-Source LLMs on open coding task across different adaptation methods and prompts. This table presents the BERTScore and ROUGE scores for each model, indicating precision (\textit{P}), recall (\textit{R}), and F1 scores for BERTScore, along with ROUGE scores (1, 2, L). Models were evaluated under different scenarios, including finetuning and few-shot approaches, with prompts designed to align with thematic analysis. }

    \label{tab:main-results}
\end{table*}


Upon analysis, we found that the averaged difficulty metric correlated strongly with the length of the sentences. Longer sentences tended to be perceived as more complex. In contrast, traditional lexicon-based readability metrics, such as Flesch Reading Ease \cite{kincaid1975derivation} and the Coleman-Liau Index \cite{coleman1975computer}\footnote{Ward, Alex. 2022. Textstat, \url{https://pypi.org/project/textstat/}}, were found to be uncorrelated with the difficulty ratings assigned by the coders. These readability scores, designed for general text comprehension, failed to capture the specific challenges associated with qualitative coding (see Appendix C). As a result, sentence length and the coders' averaged perceived difficulty were more reliable indicators of complexity in this open coding task.

\paragraph{Stage 2: Rating Coder and Model Labels}
In the second stage, the coders were provided with the labels generated by the other coders, the labels generated by the best-performing LLM models from the first stage, as well as golden standard labels. The coders were asked to rate each label on a scale from 1 to 5, with 5 representing the most accurate and representative coding of the sentence's core idea. This stage aimed to evaluate both the quality of human-generated codes and the performance of the LLMs in comparison to them. The instructions for coders are attached in Appendix E. Two key metrics were used in the evaluation:

\begin{figure}
    \centering
    \includegraphics[width=0.8\linewidth]{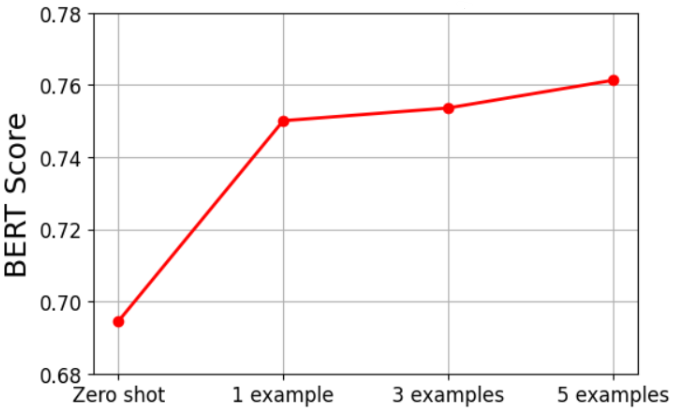}
    \caption{Mistral BERT F1 scores across different numbers of examples.}
    \label{fig:fewshot}
\end{figure}

\subsection{Metrics}

\paragraph{Deviation from golden standard} For each coder \( i \) and each sentence \( j \), the deviation was calculated by comparing the average rating of the humans' and LLMs' codes with the golden standard label. The deviation for each code was computed using the formula: 

\[
\text{DGS}_{i,j} = \left( \frac{1}{N} \sum_{k=1}^{N} r_{k,j}^{(i)} \right) - \left( \frac{1}{N} \sum_{k=1}^{N} r_{k,j}^{(\text{GS})} \right)
\]

where:\\
- \( i \) is the coder/model for whom the deviation is calculated. \\
- \( j \) is the specific sentence being evaluated.\\
- \( N \) represents the number of experts who rated both the coder/model \( i \) and the golden standard (GS) for sentence \( j \).\\
- \( r_{k,j}^{(i)} \) is the rating given by expert \( k \) to coder/model \( i \) for sentence \( j \).\\
- \( r_{k,j}^{(\text{GS})} \) is the rating given by expert \( k \) to the golden standard (GS) for sentence \( j \).

\paragraph{Average DGS} 
To compute an overall measure of deviation for each coder \( i \) across all sentences (\( M \) - total number of sentences), the average deviation was calculated as follows:

\[
\text{Average DGS}_{i} = \frac{1}{M} \sum_{j=1}^{M} \text{DGS}_{i,j}
\]

In this case, \textit{positive deviation} occurs when an expert rates a code higher than the golden standard, resulting in a positive deviation from it. \textit{Negative deviation}, on the other hand, is when a coder rates a sentence lower than the golden standard. Both types indicate a coder's divergence from the golden standard but in different directions, reflecting higher or lower evaluation by humans of a particular code.\\    



\paragraph{Inter-Coder Reliability} To assess the reliability and consistency of the codes generated by different coders, including both human coders and LLMs, Krippendorff's alpha \cite{krippendorff2018content} was computed. Krippendorff's alpha is a widely used reliability coefficient that quantifies the level of agreement between coders on a set of coding tasks, while accounting for the possibility of chance agreement. It is particularly versatile, as it can handle various types of data, including nominal, ordinal, interval, and ratio-level data, making it well-suited for qualitative research where different coding schemes or scales are used.

Krippendorff’s alpha is valuable because it accommodates situations where coders may not agree perfectly and where missing or incomplete data is present, unlike simpler agreement measures like percent agreement or Cohen’s kappa \cite{mchugh2012interrater}, which require complete data and assume equal distribution across categories. It can also handle any number of coders, not just pairs, making it ideal for our study, which involves multiple human coders and LLMs.



\subsection{Results}

\begin{figure*}[ht]
    \centering
    \subfigure[Average rating of LLMs and Humans]{
        \includegraphics[width=0.47\linewidth]{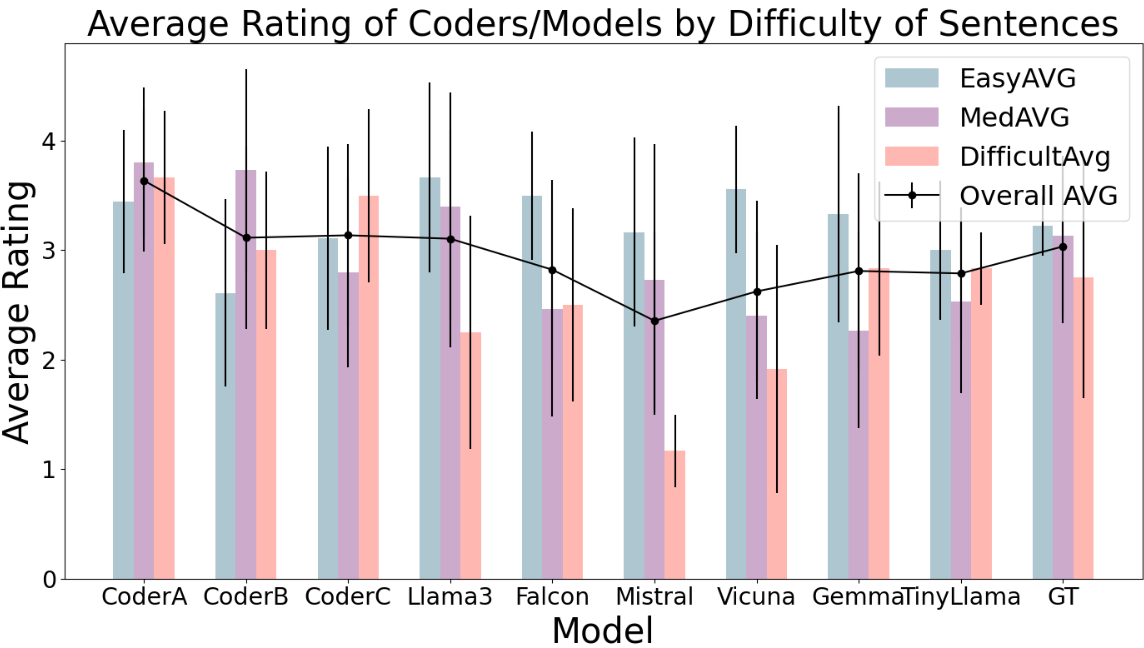}
        \label{fig:ratingavg}
    }
    \subfigure[Deviation evaluation]{
        \includegraphics[width=0.48\linewidth]{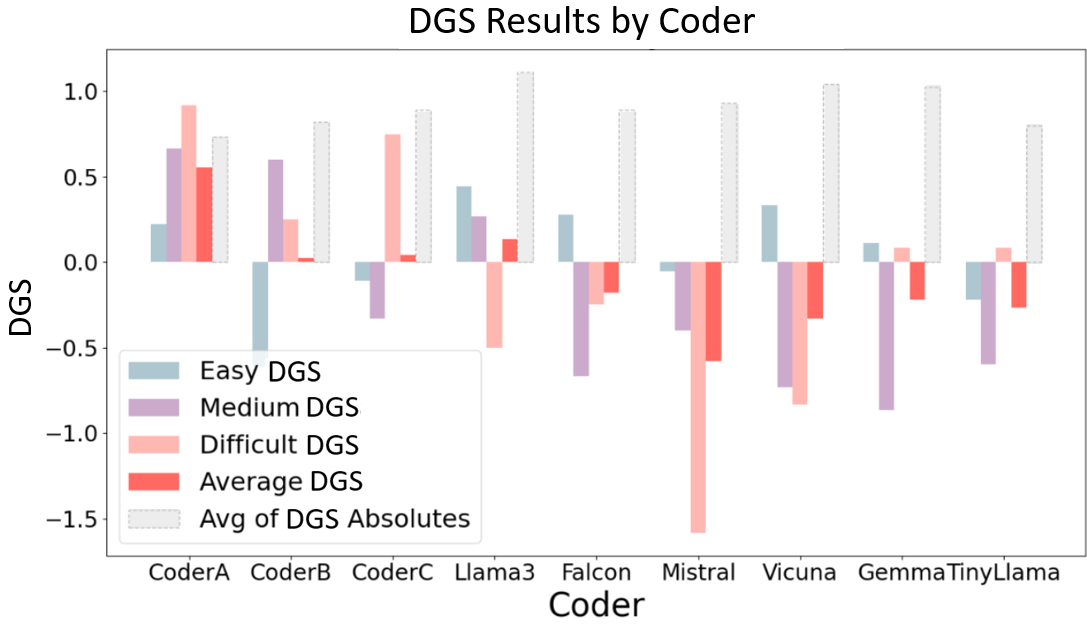}
        \label{fig:biases}
    }
    \caption{Comparison of Average Ratings and Deviation from Golden Standard (DGS) for LLMs and human coders. Panel (a) shows the average ratings given to both human coders (CoderA, CoderB, CoderC) and various LLM models, segmented by sentence difficulty (Easy, Medium, Difficult). The graph highlights that LLMs generally receive higher ratings on easy sentences compared to human coders, while humans excel in coding more complex sentences. Panel (b) presents the DGS results for both human coders and LLMs across different sentence difficulties, with positive and negative deviations from the golden standard.}
    \label{fig:combined}
\end{figure*}

The result shown in the Figure \ref{fig:ratingavg} indicate that human coders performed exceptionally well in coding difficult sentences, which often involved abstract concepts or nuanced language. However, their performance was less consistent with easier sentences, where LLMs tended to perform better. This discrepancy is likely due to the tendency of human coders to overcomplicate simple statements or overlook straightforward interpretations. One coder, in particular, commented during the evaluation phase that they tended to overinterpret data and make codes too abstract.

This tendency was shown when coding simple sentences, for instance \textit{I can ask the voice assistant what the weather is like.} In this example, LLMs generated codes such as \textit{weather forecast} or \textit{weather prediction} which are similar to Golden Standard label \textit{weather}. However, humans provided more abstract code: \textit{functional usage, device feature,} and \textit{voice command.} While these additional layers of interpretation may add depth in some contexts, in this case, they introduced unnecessary complexity and deviated from the core meaning of the sentence. Nevertheless, this level of abstraction could be valuable for the next stage of axial coding where codes are organized into hierarchies.


This tendency of human coders to overcomplicate simple sentences was also reflected in the DGS evaluation results (see Figure \ref{fig:biases}). For human coders (Coders A, B, and C), deviation from golden standard was particularly pronounced for easier sentences. Coder A, in particular, consistently showed positive deviation, being further from golden standard but rated higher by experts. In contrast, LLMs generally exhibited less deviation across sentence complexities. For instance, Llama3 demonstrated positive deviation for medium and difficult sentences, suggesting that it tended to overpredict or generate overly complex codes in certain cases that mimics human expert behavior. Models like Falcon and Mistral showed much lower deviation, particularly for easy and medium sentences, where their labels aligned more closely with the golden standard. Overall, LLMs demonstrated lower and more consistent deviation compared to human coders, particularly for easier sentences. This suggests that LLMs are more reliable in handling straightforward coding tasks. However, as the sentence complexity increased, some models, such as Llama3, exhibited positive deviation, meaning being evaluated higher than golden standard by experts, while other LLMs showed the opposite trend. In contrast, human coders, while showing higher deviation overall, were able to better handle the complexity of difficult sentences, albeit inconsistently.


Despite the effort to ensure consistency in the coding process, the inter-coder reliability, measured using Krippendorff's alpha, was low, with a value of 0.2. This low value indicates a significant lack of agreement between coders, which can be attributed to the subjective nature of the task and the inherent variability in how individuals interpret complex, abstract concepts \cite{hayes2007answering}. Additionally, the broad definition of a code provided by \citet{Saldana2016} may have allowed for considerable variation in how coders applied and interpreted the codes, further contributing to the low reliability. In comparison, tasks like rating restaurant experiences or product reviews may be better suited to this evaluation metric because they involve more objective criteria (e.g., food quality, and service speed). The task of qualitative code evaluation, however, involves a higher degree of interpretation and abstraction \cite{galdas2017revisiting}, making it less suitable for standard reliability metrics like Krippendorff's alpha.



\section{Discussion}

The results of this study revealed several surprising and, at times, counterintuitive findings. Notably, BERTScore performance plateaued after approximately 100 examples, suggesting that effective fine-tuning is achievable with a relatively small dataset. This has important implications for computational social science research, where data is often scarce and difficult to collect.

One of the most unexpected outcomes was that LLMs exhibited less deviation in coding, meaning their outputs were often closer to the golden standard compared to human coders. This finding challenges the common assumption that human coders, with their deep contextual understanding and expertise, would naturally generate more accurate and reliable codes. 

Human coders had a tendency to overinterpret the data, adding unnecessary complexity to straightforward sentences. As it was highlighted by one of the experts, this is a pitfall in early qualitative analysis and may reflect an effort to capture nuances that aren't immediately relevant in the initial open coding phase. Interestingly, as research progresses into more advanced stages such as axial coding, these preliminary codes are often refined and simplified. Therefore, it would be fascinating to investigate how human coders' open codes might change if they were later exposed to the results of axial coding (higher order categories).

\section{Conclusion}

For the task of open coding, we tested six open-source LLMs and compared their performance in finetuning, zero-shot, and few-shot scenarios. Following this, we conducted a human expert evaluation to compare the codes produced by LLMs with those created by human coders. This comparative approach allowed us to assess the strengths and limitations of LLMs in automating qualitative coding tasks.

The study revealed important insights into the potential of LLMs, as well as the challenges they face. While for short sentences LLMs tend to be closer to golden standard labels than human coders, they lack the interpretative depth necessary for complex qualitative analysis. Human coders, despite their expertise, often introduce unnecessary complexity into their codes, reflecting a tendency to overinterpret data during the open coding phase. The results demonstrated that LLMs hold great promise in automating the open coding process, especially in domains where the data is straightforward and repetitive, such as customer feedback analysis or social media monitoring. However, for more nuanced tasks, particularly in academic social science research, human coders remain more reliable due to their ability to interpret longer and more difficult sentences that LLMs struggle to handle.

Future work will extend this study into the axial coding stage, with the goal of developing a complete thematic analysis pipeline and evaluating its performance against human expert results. This next phase will assess how effectively LLMs can contribute to the full qualitative coding process and determine whether they are suitable for full automation or better suited as an assistive tool.

\section*{Ethics Statement}

The use of LLMs in qualitative research introduces new ethical considerations, particularly concerning bias and the potential for automated systems to replicate or amplify human biases. In this study, we took steps to identify and mitigate biases in the models and human coders. However, we acknowledge that the use of LLMs in sensitive research areas requires the development of guidelines to ensure that these tools are used responsibly. The human participants involved in the expert evaluation were fully informed about the study's objectives and provided their consent to participate. Their expertise was crucial in evaluating the performance of the LLMs, and their input was treated with the utmost respect and consideration.

\section*{Limitations}

Firstly, it is important to highlight that our focus was limited to open coding; we did not explore the full qualitative analysis process, particularly axial coding, which organizes open codes into higher-order categories. Future research could extend this work by investigating whether LLMs can assist in axial coding, potentially offering a complete automation of thematic analysis.

The dataset used in this study focused primarily on social science research, supplemented by online reviews. However, the scope of QDA extends beyond this, encompassing domains such as social media posts, medical texts, media content, and field notes. It means that the current models may not generalize well across all potential domains, especially those with specialized terminology and professional knowledge requirements.

Another limitation is that while we used established metrics such as BERTScore and ROUGE, these may not fully capture the quality and interpretative nature of qualitative coding. Developing more nuanced evaluation metrics that align better with the goals of qualitative research would be an important step forward.

Lastly, this study compared LLM performance to human coders based on alignment with a golden standard, which may not be the ideal measure of codes' quality.  In real-world coding scenarios, human coders often reach a consensus after discussion, whereas LLMs do not undergo such collaborative process. This raises the question of whether LLM-generated codes could eventually reach consensus between models or if their role is better suited as an assistive tool for human coders. Further investigation into a hybrid approach—where LLMs handle initial coding and humans provide further refinement and interpretation, especially for complex or ambiguous data—would be a valuable direction for future research.

\bibliography{acl_latex}

\appendix

\section{Prompts used for finetuning}

In the conducted experiments, several prompts were designed to guide the models in performing open coding tasks. These prompts varied in their level of explicitness, the perspective they asked the model to adopt, and the amount of detail they requested. Additionally, the experiments compared the effect of using a line break versus a period (dot) at the end of each prompt to assess how subtle changes in prompt formatting might influence the model's performance (see Appendix B). Below is a brief description of each prompt:

\begin{itemize}
    \item \textbf{Explicit Instruction (Prompt 1):} \textit{Summarize the main idea of a sentence.} This prompt provides a direct and clear instruction to the model, asking it to summarize the core idea of a given sentence. The expectation is for the model to extract the primary message or theme conveyed in the sentence with no additional context or framing. This prompt is designed to test the model's ability to perform a straightforward task without needing implicit knowledge.

    \item \textbf{Informal Request (Prompt 2):} \textit{Can you tell me what the main idea of this sentence is in just a few words?} This prompt is phrased as a casual, conversational question, asking the model to summarize the sentence in "just a few words." The informal tone encourages a more concise and simplified response, aiming to capture how well the model can extract the essence of the sentence in a more natural, everyday context. 

    \item \textbf{Expert Angle (Prompt 3):} \textit{From the perspective of a social scientist, summarize the following sentence as you would in thematic coding.} This prompt takes a more specialized approach, asking the model to assume the perspective of a social scientist performing thematic coding. The expectation here is for the model to not only summarize the sentence but to apply a more analytical and structured lens, possibly introducing higher-level categorizations that would be typical in qualitative data analysis. 

    \item \textbf{Impersonalization (Prompt 4):} \textit{If you were a social scientist doing thematic analysis, what code would you give to this citation?} In this prompt, the model is asked to act as a social scientist and assign a code, which is a brief label representing the central idea of the sentence. It emphasizes the objectivity of thematic analysis, expecting the model to depersonalize the task and focus on generating an appropriate label that accurately reflects the content. 

    \item \textbf{Detailed Explanation (Prompt 5):} \textit{Explain in a couple of words the primary thought expressed in the following text.} This prompt asks the model to provide a more detailed, thorough explanation of the primary thought behind the text. It is designed to encourage the model to go beyond a simple summary and delve into the deeper meaning or implications of the sentence. 

    \item \textbf{Simplified Task (Prompt 6):} \textit{What is the gist of this sentence?} This prompt simplifies the task by asking for the "gist" of the sentence. It challenges the model to provide a very brief and straightforward summary, focusing on distilling the essential meaning of the sentence. 
\end{itemize}

\section{Detailed fine-tuning results}

These results (see Table \ref{tab:detailed-finetuning-results}) demonstrate the performance of various models when fine-tuned on the task of open coding using different prompts. BERTScore and ROUGE are reported.

\begin{table*}[]
\centering
\tiny

\resizebox{\linewidth}{!}{
\begin{tabularx}{\linewidth}{>{\raggedright\arraybackslash}X >{\centering\arraybackslash}X >{\centering\arraybackslash}X >{\centering\arraybackslash}X >{\centering\arraybackslash}X >{\centering\arraybackslash}X >{\centering\arraybackslash}X}
    \toprule
    \textbf{Model} & \multicolumn{3}{c}{\textbf{BERTScore}} & \multicolumn{3}{c}{\textbf{ROUGE}} \\ 
    \cmidrule(lr){2-4} \cmidrule(lr){5-7}
    & \textbf{$P_{std}$} & \textbf{$R_{std}$} & \textbf{$F1_{std}$} & \textbf{1} & \textbf{2} & \textbf{L} \\
    \midrule

    \multicolumn{7}{c}{\textbf{Summarize the main idea of a sentence$\backslash$n}} \\ 
    \midrule
    Llama3    & $0.713_{0.060}$ & $0.758_{0.040}$ & $0.734_{0.062}$ & $0.141$ & $0.033$ & $0.153$ \\
    Falcon    & $0.746_{0.073}$ & $0.782_{0.097}$ & $0.764_{0.095}$ & $0.176$ & $0.047$ & $0.189$ \\
    Mistral   & $0.729_{0.076}$ & $0.787_{0.093}$ & $0.756_{0.078}$ & $0.178$ & $0.047$ & $0.195$ \\
    Vicuna    & $0.731_{0.063}$ & $0.777_{0.095}$ & $0.753_{0.079}$ & $0.163$ & $0.028$ & $0.182$ \\
    Gemma     & $0.712_{0.084}$ & $0.738_{0.078}$ & $0.745_{0.080}$ & $0.163$ & $0.030$ & $0.168$ \\
    TinyLlama & $0.718_{0.072}$ & $0.775_{0.090}$ & $0.757_{0.087}$ & $0.164$ & $0.052$ & $0.158$ \\

    \midrule

    \multicolumn{7}{c}{\textbf{Summarize the main idea of a sentence.}} \\ 
    \midrule
    Llama3    & $0.718_{0.072}$ & $0.788_{0.089}$ & $0.750_{0.073}$ & $0.181$ & $0.059$ & $0.166$ \\
    Falcon    & $0.738_{0.099}$ & $0.787_{0.103}$ & $0.761_{0.096}$ & $0.213$ & $0.077$ & $0.210$ \\
    Mistral   & $0.719_{0.072}$ & $0.768_{0.086}$ & $0.742_{0.075}$ & $0.157$ & $0.055$ & $0.148$ \\
    Vicuna    & $0.733_{0.079}$ & $0.787_{0.095}$ & $0.758_{0.081}$ & $0.193$ & $0.068$ & $0.185$ \\
    Gemma     & $0.719_{0.071}$ & $0.779_{0.089}$ & $0.746_{0.072}$ & $0.172$ & $0.049$ & $0.166$ \\
    TinyLlama & $0.736_{0.083}$ & $0.788_{0.092}$ & $0.760_{0.081}$ & $0.207$ & $0.074$ & $0.199$ \\
    \midrule

    \multicolumn{7}{c}{\textbf{Can you tell me what the main idea of this sentence is in just a few words?}} \\ 
    \midrule
    Llama3    & $0.688_{0.055}$ & $0.778_{0.084}$ & $0.729_{0.061}$ & $0.116$ & $0.034$ & $0.110$ \\
    Falcon    & $0.753_{0.105}$ & $0.787_{0.108}$ & $0.768_{0.102}$ & $0.236$ & $0.104$ & $0.239$ \\
    Mistral   & $0.742_{0.106}$ & $0.795_{0.106}$ & $0.766_{0.101}$ & $0.246$ & $0.106$ & $0.235$ \\
    Vicuna    & $0.691_{0.060}$ & $0.783_{0.087}$ & $0.732_{0.063}$ & $0.168$ & $0.047$ & $0.164$ \\
    Gemma     & $0.711_{0.075}$ & $0.786_{0.093}$ & $0.746_{0.079}$ & $0.171$ & $0.057$ & $0.168$ \\
    TinyLlama & $0.725_{0.083}$ & $0.789_{0.090}$ & $0.754_{0.079}$ & $0.178$ & $0.067$ & $0.177$ \\
    \midrule

    \multicolumn{7}{c}{\textbf{From the perspective of a social scientist, summarize the following sentence as you would in thematic coding$\backslash$n}} \\ \midrule
    Llama3    & $0.698_{0.059}$ & $0.784_{0.083}$ & $0.738_{0.062}$ & $0.130$ & $0.033$ & $0.119$ \\
    Falcon    & $0.745_{0.109}$ & $0.792_{0.105}$ & $0.766_{0.102}$ & $0.210$ & $0.089$ & $0.211$ \\
    Mistral   & $0.688_{0.060}$ & $0.785_{0.086}$ & $0.732_{0.064}$ & $0.139$ & $0.041$ & $0.131$ \\
    Vicuna    & $0.713_{0.080}$ & $0.778_{0.094}$ & $0.743_{0.080}$ & $0.169$ & $0.061$ & $0.166$ \\
    Gemma     & $0.721_{0.085}$ & $0.784_{0.093}$ & $0.749_{0.082}$ & $0.180$ & $0.070$ & $0.177$ \\
    Tinyllama & $0.718_{0.073}$ & $0.776_{0.083}$ & $0.745_{0.072}$ & $0.165$ & $0.053$ & $0.158$ \\

    \midrule
    
    \multicolumn{7}{c}{\textbf{From the perspective of a social scientist, summarize the following sentence as you would in thematic coding.}} \\ \midrule

    Llama3    & $0.685_{0.082}$ & $0.781_{0.064}$ & $0.733_{0.081}$ & $0.136$ & $0.025$ & $0.154$ \\
    Falcon    & $0.754_{0.066}$ & $0.778_{0.091}$ & $0.759_{0.088}$ & $0.181$ & $0.048$ & $0.190$ \\
    Mistral   & $0.740_{0.067}$ & $0.780_{0.088}$ & $0.756_{0.071}$ & $0.172$ & $0.045$ & $0.187$ \\
    Vicuna    & $0.718_{0.071}$ & $0.780_{0.094}$ & $0.753_{0.073}$ & $0.165$ & $0.039$ & $0.185$ \\
    Gemma     & $0.700_{0.072}$ & $0.780_{0.085}$ & $0.746_{0.080}$ & $0.180$ & $0.046$ & $0.187$ \\
    TinyLlama & $0.729_{0.076}$ & $0.778_{0.089}$ & $0.754_{0.080}$ & $0.169$ & $0.046$ & $0.183$ \\

    \midrule
    
    \multicolumn{7}{c}{\textbf{ If you were a social scientist doing thematic analysis, what code would you give to this citation?}} \\ \hline
    Llama3    & $0.692_{0.060}$ & $0.785_{0.083}$ & $0.735_{0.064}$ & $0.064$ & $0.043$ & $0.126$ \\
    Falcon    & $0.736_{0.093}$ & $0.785_{0.101}$ & $0.759_{0.092}$ & $0.206$ & $0.076$ & $0.200$ \\
    Mistral   & $0.686_{0.057}$ & $0.785_{0.082}$ & $0.731_{0.061}$ & $0.132$ & $0.044$ & $0.123$ \\
    Vicuna    & $0.719_{0.070}$ & $0.789_{0.091}$ & $0.751_{0.073}$ & $0.183$ & $0.063$ & $0.169$ \\
    Gemma     & $0.724_{0.085}$ & $0.784_{0.091}$ & $0.751_{0.082}$ & $0.170$ & $0.066$ & $0.168$ \\
    Tinyllama & $0.720_{0.071}$ & $0.778_{0.083}$ & $0.747_{0.072}$ & $0.186$ & $0.053$ & $0.182$ \\

    \midrule

    \multicolumn{7}{c}{\textbf{What is the gist of this sentence?}} \\ \midrule
            
    Llama3    & $0.680_{0.064}$ & $0.780_{0.086}$ & $0.725_{0.066}$ & $0.129$ & $0.042$ & $0.121$ \\
    Falcon    & $0.731_{0.091}$ & $0.780_{0.098}$ & $0.754_{0.089}$ & $0.182$ & $0.080$ & $0.179$ \\
    Mistral   & $0.726_{0.079}$ & $0.785_{0.095}$ & $0.753_{0.079}$ & $0.165$ & $0.057$ & $0.160$ \\
    Vicuna    & $0.720_{0.070}$ & $0.781_{0.089}$ & $0.748_{0.072}$ & $0.172$ & $0.055$ & $0.162$ \\
    Gemma     & $0.707_{0.077}$ & $0.773_{0.091}$ & $0.737_{0.076}$ & $0.152$ & $0.059$ & $0.146$ \\
    Tinyllama & $0.713_{0.057}$ & $0.773_{0.079}$ & $0.741_{0.061}$ & $0.143$ & $0.032$ & $0.139$ \\

    \midrule

    \multicolumn{7}{c}{\textbf{Explain in a couple of words the primary thought expressed in the following text$\backslash$n}} \\ \midrule
            
    Llama3    & $0.691_{0.062}$ & $0.783_{0.085}$ & $0.733_{0.066}$ & $0.120$ & $0.038$ & $0.110$ \\
    Falcon    & $0.734_{0.078}$ & $0.778_{0.090}$ & $0.754_{0.078}$ & $0.171$ & $0.049$ & $0.165$ \\
    Mistral   & $0.698_{0.067}$ & $0.780_{0.088}$ & $0.735_{0.070}$ & $0.141$ & $0.038$ & $0.131$ \\
    Vicuna    & $0.703_{0.072}$ & $0.780_{0.088}$ & $0.738_{0.072}$ & $0.155$ & $0.048$ & $0.148$ \\
    Gemma     & $0.706_{0.064}$ & $0.786_{0.086}$ & $0.742_{0.066}$ & $0.177$ & $0.053$ & $0.170$ \\
    Tinyllama & $0.720_{0.077}$ & $0.784_{0.091}$ & $0.750_{0.078}$ & $0.168$ & $0.071$ & $0.163$ \\

    \midrule

    \multicolumn{7}{c}{\textbf{Explain in a couple of words the primary thought expressed in the following text.}} \\ \midrule
    Llama3    & $0.700_{0.068}$ & $0.784_{0.055}$ & $0.747_{0.063}$ & $0.142$ & $0.025$ & $0.152$ \\
    Falcon    & $0.752_{0.088}$ & $0.779_{0.061}$ & $0.760_{0.086}$ & $0.183$ & $0.042$ & $0.193$ \\
    Mistral   & $0.738_{0.070}$ & $0.790_{0.090}$ & $0.759_{0.073}$ & $0.173$ & $0.047$ & $0.183$ \\
    Vicuna    & $0.717_{0.066}$ & $0.780_{0.094}$ & $0.752_{0.099}$ & $0.161$ & $0.025$ & $0.182$ \\
    Gemma     & $0.708_{0.068}$ & $0.778_{0.079}$ & $0.746_{0.098}$ & $0.172$ & $0.039$ & $0.186$ \\
    TinyLlama & $0.728_{0.073}$ & $0.778_{0.091}$ & $0.755_{0.089}$ & $0.168$ & $0.053$ & $0.168$ \\

    \bottomrule
\end{tabularx}

}
\caption{Detailed Fine-tuning Results. The following table presents the detailed results from fine-tuning experiments, including precision (P), recall (R), F1 score, and ROUGE across different models and prompts.}
\label{tab:detailed-finetuning-results}
\end{table*}

\section{Sentence difficulty and lexicon-based metrics}

In this section we present the Table \ref{tab:difficulty_metrics} with the assessed difficulty levels of sentences using a range of lexicon-based metrics. This includes readability scores like Flesch Reading Ease and Coleman-Liau Index, along with indicators of linguistic complexity such as sentence length and syllable count. The table provides a color-coded overview of sentence difficulty based on calculated average perceived difficulty of a sentence.

\begin{table*}[]
\centering
\tiny
\begin{tabular}{|p{2cm}|c|c|c|c|c|c|c|c|c|c|}
\hline

\textbf{Sentence}                                                                                                 & \textbf{Length} & \textbf{Difficulty} & \textbf{Difficulty} & \textbf{Flesch Reading} & \textbf{Coleman-} & \textbf{Automated} & \textbf{Difficult} & \textbf{Syllable}  \\ 
                                                                                                                  &                 & \textbf{assumed}    & \textbf{avg}        & \textbf{Ease}           & \textbf{Liau}    & \textbf{Index}     & \textbf{Words}     & \textbf{count}  \\ \hline

Quote1                                                                & 51.00           & 1.00                       & \cellcolor[HTML]{D9EAD3}1.50                & 46.44                     & 10.80                     & 8.50                      & 1.00                         & 14.00                                       \\ \hline
Quote2                                                             & 55.00           & 1.00                       & \cellcolor[HTML]{D9EAD3}1.50                & 85.69                     & 4.74                      & 3.30                      & 1.00                         & 14.00                                      \\ \hline
Quote3                                          & 75.00           & 1.00                       & \cellcolor[HTML]{D9EAD3}1.00                & 42.38                     & 12.28                     & 9.70                      & 4.00                         & 21.00                                      \\ \hline
Quote4              & 101.00          & 2.00                       & \cellcolor[HTML]{D9EAD3}1.50                & 103.12                    & 3.05                      & 2.20                      & 2.00                         & 23.00                                       \\ \hline

Quote5          & 105.00          & 1.00                       & \cellcolor[HTML]{D9EAD3}1.50                & 87.72                     & 8.03                      & 6.30                      & 3.00                         & 24.00                                 \\ \hline

Quote6 & 155.00          & 1.00                       & \cellcolor[HTML]{D9EAD3}1.50                & 67.42                     & 7.50                      & 13.70                     & 3.00                         & 39.00                                                \\ \hline
Quote7 & 179.00          & 3.00                       & \cellcolor[HTML]{FFF2CC}2.25                & 88.74                     & 4.83                      & 4.00                      & 4.00                         & 42.00                                         \\ \hline
Quote8 & 194.00          & 2.00                       & \cellcolor[HTML]{FFF2CC}1.75                & 40.01                     & 13.41                     & 19.00                     & 6.00                         & 51.00                                                \\ \hline
Quote9 & 289.00          & 2.00                       & \cellcolor[HTML]{FFF2CC}2.00                & 91.82                     & 5.79                      & 6.10                      & 8.00                         & 65.00                                      \\ \hline
Quote10 & 291.00          & 2.00                       & \cellcolor[HTML]{FFF2CC}1.75                & 85.89                     & 4.39                      & 4.60                      & 5.00                         & 70.00                                     \\ \hline
Quote11 & 309.00          & 3.00                       & \cellcolor[HTML]{FFF2CC}2.25                & 80.17                     & 4.54                      & 3.50                      & 7.00                         & 77.00                                          \\ \hline
Quote12 & 350.00          & 3.00                       & \cellcolor[HTML]{FCE5CD}2.75                & 67.79                     & 9.05                      & 11.10                     & 9.00                         & 86.00                                     \\ \hline
Quote13 & 379.00          & 2.00                       & \cellcolor[HTML]{FCE5CD}2.75                & 72.19                     & 6.22                      & 10.60                     & 6.00                         & 96.00                                              \\ \hline
Quote14 & 889.00          & 3.00                       & \cellcolor[HTML]{FCE5CD}2.75                & 75.20                     & 7.53                      & 7.20                      & 21.00                        & 218.00                                       \\ \hline
Quote15 & 1117.00          & 3.00                       & \cellcolor[HTML]{FCE5CD}2.75                & 54.09                     & 8.43                      & 16.70                     & 22.00                        & 282.00                                           \\ \hline
\end{tabular}
\caption{Sentence Difficulty and Lexicon-based Metrics. Full quotes are in Appendix E.}
\label{tab:difficulty_metrics}
\end{table*}

\section{Sentences used for expert coding}

This section lists the specific sentences that were selected from the test set for the expert coding phase (see Table \ref{tab:expert_coding_sentences}). These sentences span a broad spectrum of themes and linguistic features, from simple descriptive statements to complex and long sentences.

\begin{table*}[h!]
    \centering
    \tiny
    \begin{tabular}{c p{12cm}}
        \toprule
        \textbf{Num} & \textbf{Sentence} \\ 
        \midrule
        Quote1 & She doesn't always understand correctly what I say. \\ 
        Quote2 & I can ask the voice assistant what the weather is like. \\ 
        Quote3 & The food was delicious and the waiter was incredibly helpful and attentive. \\ 
        Quote4 & I was so worried. The thoughts kept spinning in my head and I’d lay there with my eyes open for hours. \\ 
        Quote5 & Had a great experience at this restaurant... staff was pleasant; food was tasty and large in portion size. \\ 
        Quote6 & I wasn't really as concerned about portability (it's a very large laptop) but it's not hard to move around or take on a trip which was a pleasant surprise. \\ 
        Quote7 & Well, often people eat up guilt. They start to, I don't know, do bad things. They start drinking, for example. I don't know, they start to ruin themselves in every possible way. \\ 
        Quote8 & I think that everything is possible in this world, that everyone will definitely understand that there is happiness for them, so it gives them positive emotions and develops their values within. \\ 
        Quote9 & ...there are a lot of memes about Duolingo, even if you look on the Internet. About this green owl, which, if you haven't started speaking Mexican, Spanish, comes to you and kills your whole family with a shotgun. It's very active there, yes. As if you should, you promised us and so on. \\ 
        Quote10 & Interviewer: Okay, can procrastination be funny? Informant: Not funny, THIS IS NOT FUNNY. I don't know why do I do this... you know... it's like... feeling like you don't want to learn. It's not funny because it requires rest, you can't get it out of your head that you don't want to do it. \\ 
        Quote11 & Well, there is a probability. But I would say that each person has his own head on his shoulders. That is, he defines his own barriers, as they say. Every person determines his own barriers. I mean, if he's, I don't know. If he wants to take risks, let him take risks. It's everybody's business in principle. \\ 
        Quote12 & For me, fear is a constant companion. I think, because shame and fear, well, shame is probably an emotion that you realize that you are justifiably ashamed, probably, that is certain moral norms that you have violated, so you are ashamed. Fear, on the other hand, is an emotion that can occur regardless of whether you've done something wrong or not. \\ 
        Quote13 & I don't sleep much, only 5 hours a day. But I know sleep is important, because I am often tired and, due to lack of sleep, I cannot listen. It's probably important to go to classes, rest and then start studying again, although I have a lot of problems with this, because if I rest, I don't start studying again, so for me it's better to continue studying during the day to finish. \\ 
        Quote14 & House in some small town, or at least I live in Korolev, in this area. Pets and everything connected with it. I would like to see this come true. Some kind of stability, a well-paid job. Positive feelings... in some kind of future, not quite distant, but not exactly near, which is exactly tomorrow. Can plan for a couple of months (it all depends on how my goals in life change, it's all very changeable, it seems to me). Ten years later - still a husband, perhaps children, but not a fact. Stable work is quite possible. Animals are possible, maybe not in the house, but in the apartment. If you earn enough money, you can have a separate apartment. The only obstacle that can become is myself, some complexes, self-doubt, perhaps psychological problems, all that. If, for example, you don't have enough qualifications for a job, you can finish your studies and gain some more knowledge. \\ 
        Quote15 & I understand that there is an expression in the Bible that everything works out for the good of the loving God, and in my life, analyzing global situations from above and having lived some certain things in my life, milestones, I saw that having made this or that choice, having destroyed something, having left somewhere, for example, having changed even my job, initially I didn't want to lose it, I didn't want to leave this situation. I was comfortable there, it was good for me, and suddenly it all broke, destroyed, I think ah how bad, ah how sad, but having experienced all this and looking back, I realize that I gained much more. That is, I have morally matured, I have experienced some things, I realized that I will know in the next situation what to do and how to act, how to perceive. And globally I gained more, i.e. I, for example, found myself in another job, which now suits me better or at that time suited me better than what I had before. Even though initially it was just the collapse of my whole life, everything, everything is very bad there, I mean I realize that everything ends well anyway. \\ 
        \bottomrule
    \end{tabular}
    \caption{Sentences used for expert coding.}
    \label{tab:expert_coding_sentences}
\end{table*}

\section{Instructions given to coders in the second stage}

\begin{figure*}
    \centering
    \includegraphics[width=0.9\linewidth]{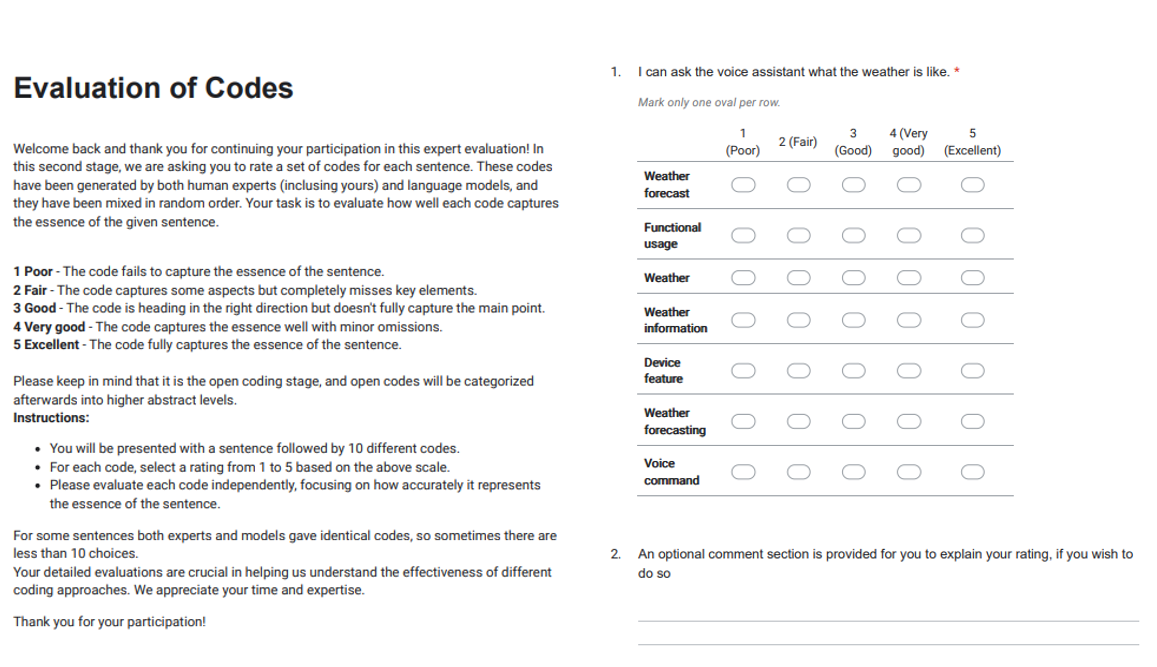}
    \caption{Screenshot of a user interface for the second stage of expert evaluation.}
    \label{fig:expert2}
\end{figure*}

This section describes the instructions provided to coders during the second stage of evaluation, illustrated through an interface screenshot (see Figure \ref{fig:expert2}).

\section{Links to Models on Hugging Face}

\begin{itemize}
    \item \textbf{Llama3}: \url{https://huggingface.co/meta-llama/Meta-Llama-3-8B-Instruct}
    \item \textbf{Falcon}: \url{https://huggingface.co/tiiuae/falcon-7b-instruct}
    \item \textbf{Mistral}: \url{https://huggingface.co/mistralai/Mistral-7B-Instruct-v0.2}
    \item \textbf{Vicuna}: \url{https://huggingface.co/lmsys/vicuna-7b-v1.5}
    \item \textbf{Gemma}: \url{https://huggingface.co/google/gemma-7b-it}
    \item \textbf{TinyLlama}: \url{https://huggingface.co/TinyLlama/TinyLlama-1.1B-Chat-v1.0}
\end{itemize}

\section{LoRA Configuration}

This section provides the LoRA (Low-Rank Adaptation) configuration used for fine-tuning the models in this study. The configuration includes the rank \( r \), the scaling factor \( \alpha \), target modules, dropout rate, bias handling, and task type. Below is the code snippet used for configuring LoRA:

\tiny
\begin{verbatim}

config = LoraConfig(
    r=16,
    lora_alpha=32,
    target_modules=["gate_proj", "up_proj", "down_proj"],
    lora_dropout=0.05,
    bias="none",
    task_type="CAUSAL_LM"
)

model = get_peft_model(model, config)
print_trainable_parameters(model)

generation_config = model.generation_config
generation_config.max_new_tokens = 15
generation_config.temperature = 0.7
generation_config.top_p = 0.7
generation_config.num_return_sequences = 1
generation_config.pad_token_id = tokenizer.eos_token_id
generation_config.eos_token_id = tokenizer.eos_token_id
\end{verbatim}

\appendix

\end{document}